\documentclass[11pt,a4paper]{article}
\usepackage{authblk}
\usepackage[hyperref]{acl2019}
\usepackage{times}
\usepackage{latexsym}
\usepackage{times,latexsym}
\usepackage{url}
\usepackage[T1]{fontenc}
\usepackage{latexsym,graphicx}
\usepackage{amsmath}
\usepackage{float}
\usepackage{enumitem}
\usepackage{multirow}
\usepackage{amssymb}
\usepackage[utf8]{inputenc}
\usepackage{url}
\usepackage{amsfonts}
\usepackage{comment}
\usepackage{multirow}
\usepackage{float}
\usepackage{array}
\usepackage{subcaption}
\usepackage{hyperref}
\usepackage[font=small,labelfont=bf]{caption}
\usepackage{comment}

\DeclareMathOperator*{\relu}{ReLU}

\renewcommand\footnotemark{}

\aclfinalcopy 
\def\aclpaperid{500} 

\title{Inter-sentence Relation Extraction with Document-level\\Graph Convolutional Neural Network}

\author[1]{Sunil Kumar Sahu}
\author[1]{Fenia Christopoulou}
\author[2,3]{Makoto Miwa}
\author[1,*]{Sophia Ananiadou\thanks{$^*$Corresponding author.}}
\affil[1]{National Centre for Text Mining, \authorcr
\textnormal{\normalsize School of Computer Science, The University of Manchester, United Kingdom}}
\affil[2]{Toyota Technological Institute, Nagoya, 468-8511, Japan}
\affil[3]{Artificial Intelligence Research Center (AIRC), \authorcr
\textnormal{\normalsize National Institute of Advanced Industrial Science and Technology (AIST), Japan}}
\affil[ ]{\tt \{sunil.sahu, efstathia.christopoulou, sophia.ananiadou\}@manchester.ac.uk}
\affil[ ]{\tt makoto-miwa@toyota-ti.ac.jp}

\date{}

\begin{document}
\maketitle
\begin{abstract}
Inter-sentence relation extraction deals with a number of complex semantic relationships in documents, which require local, non-local, syntactic and semantic dependencies.
Existing methods do not fully exploit such dependencies. 
We present a novel inter-sentence relation extraction model that builds a labelled edge graph convolutional neural network model on a document-level graph. The graph is constructed using various inter- and intra-sentence dependencies to capture local and non-local dependency information.
In order to predict the relation of an entity pair, we utilise multi-instance learning with bi-affine pairwise scoring. Experimental results show that our model achieves comparable performance to the state-of-the-art neural models on two biochemistry datasets. Our analysis shows that all the types in the graph are effective for inter-sentence relation extraction.
\end{abstract}

\section{Introduction}
Semantic relationships between named entities often span across multiple sentences.
In order to extract inter-sentence relations, most approaches utilise distant supervision to automatically generate document-level corpora~\cite{peng2017,song2018n}. Recently, \citet{Patrick2018} introduced multi-instance learning (MIL)~\cite{riedel2010,surdeanu2012} to treat multiple mentions of target entities in a document. 

\begin{figure}[t!]
    \begin{center}
        \includegraphics[width=0.35\textwidth]{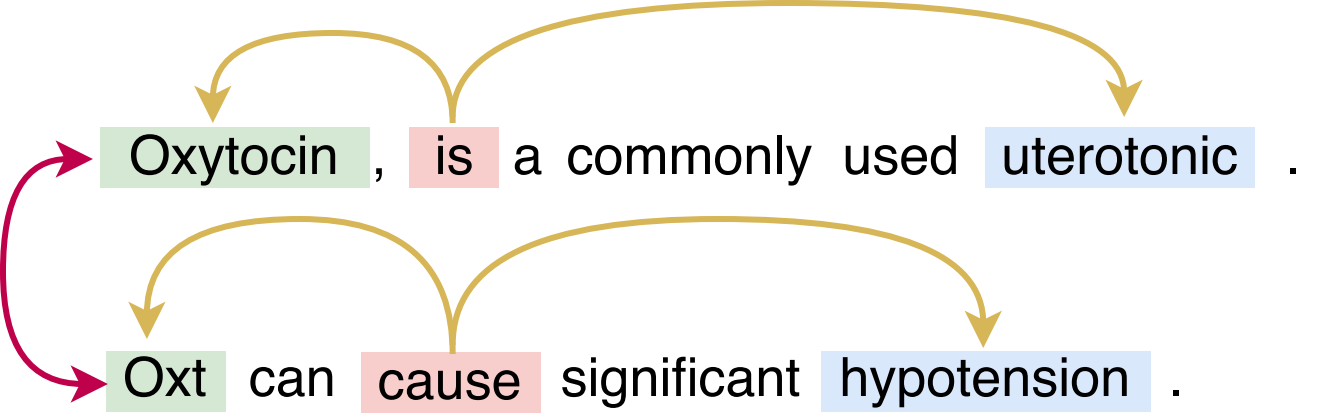}
        \caption{Sentences with non-local dependencies between named entities. The red arrow represents a relation between co-referred entities and yellow arrows represent semantically dependent relations. Example adapted from the CDR dataset~\cite{biocreative2015overview}.}
        \label{fig:graph_for_text}
    \end{center}
\end{figure}

Inter-sentential relations depend not only on local but also on non-local dependencies. Dependency trees are often used to extract local dependencies of semantic relations~\cite{culotta2004dependency,liu2015dependency} in intra-sentence relation extraction (RE). 
However, such dependencies are not adequate for inter-sentence RE, since different sentences have different dependency trees. 
Figure~\ref{fig:graph_for_text} illustrates such a case between \textit{Oxytocin} and \textit{hypotension}. 
To capture their relation, it is essential to connect the co-referring entities \textit{Oxytocin} and \textit{Oxt}.
RNNs and CNNs, which are often used for intra-sentence RE~\cite{Zeng14,dos2015,zhou2016,lin2016neural}, are not effective on longer sequences~\cite{sunil2017} thus failing to capture such non-local dependencies.

\begin{figure*}[t!]
	\begin{center}
		\includegraphics[width=0.85\textwidth]{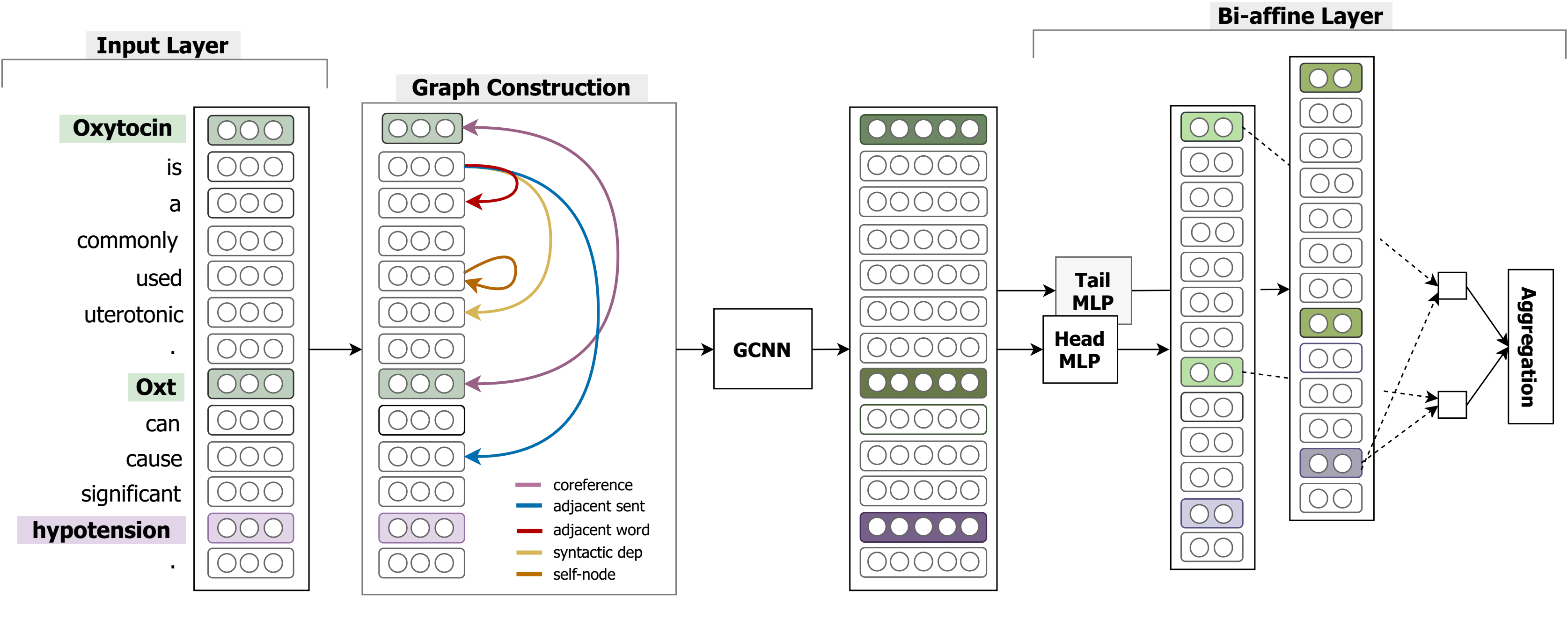}
		\caption{Proposed model architecture. The input word sequence is mapped to a graph structure, where nodes are words and edges correspond to dependencies. We omit several edges, such as self-node edges of all words and syntactic dependency edges of different labels, for brevity. GCNN is employed to encode the graph and a bi-affine layer aggregates all mention pairs.}
		\label{fig:model}
	\end{center}
\end{figure*}

We propose a novel inter-sentence RE model that builds a labelled edge Graph CNN (GCNN) model~\cite{Marcheggiani2017} on a document-level graph. 
The graph nodes correspond to words and edges represent local and non-local dependencies among them. 
The document-level graph is formed by connecting words with local dependencies from syntactic parsing and sequential information, as well as non-local dependencies from coreference resolution and other semantic dependencies~\cite{peng2017}.
We infer relations between entities using MIL-based bi-affine pairwise scoring function~\cite{Patrick2018} on the entity node representations. 

Our contribution is threefold. 
Firstly, we propose a novel model for inter-sentence RE using GCNN to capture local and non-local dependencies. 
Secondly, we apply the model on two biochemistry corpora and show its effectiveness.
Finally, we developed a novel, distantly supervised dataset with chemical reactant-product relations from PubMed abstracts.\footnote{The dataset is publicly available at \url{http://nactem.ac.uk/CHR/}.}

\section{Proposed Model}
\label{sec:proposed_model}

We formulate the inter-sentence, document-level RE task as a classification problem. 
Let $[ w_1, w_2, \cdots, w_n ]$ be the words in a document $t$ and $e_1$ and $e_2$ be the entity pair of interest in $t$. 
We name the multiple occurrences of these entities in the document \textit{entity mentions}.
A relation extraction model takes a triple ($e_1$, $e_2$, $t$) as input and returns a relation for the pair, including the ``no relation'' category, as output. 
We assume that the relationship of the target entities in $t$ can be inferred based on all their mentions. We thus apply multi-instance learning on $t$ to combine all mention-level pairs and predict the final relation category of a target pair.

We describe the architecture of our proposed model in Figure~\ref{fig:model}.
The model takes as input an entire abstract of scientific articles and two target entities with all their mentions in the input layer. 
It then constructs a graph structure with words as nodes and labelled edges that correspond to local and non-local dependencies.
Next, it encodes the graph structure using a stacked GCNN layer and classifies the relation between the target entities by applying MIL~\cite{Patrick2018} to aggregate all mention pair representations.

\subsection{Input Layer}
In the input layer, we map each word $i$ and its relative positions to the first and second target entities into real-valued vectors, $\mathbf{w}_i$, $\mathbf{d}_i^1$, $\mathbf{d}_i^2$, respectively.
As entities can have more than one mention, we calculate the relative position of a word from the closest target entity mention. 
For each word $i$, we concatenate the word and position representations into an input representation, $\mathbf{x}_i = [ \mathbf{w}_i ; \mathbf{d}_i^1  ; \mathbf{d}_i^2 ]$.

\subsection{Graph Construction}
\label{sec:graph_construction}
In order to build a document-level graph for an entire abstract, we use the following categories of inter- and intra-sentence dependency edges, as shown with different colours in Figure~\ref{fig:model}.

\noindent \textbf{Syntactic dependency edge:} 
The syntactic structure of a sentence reveals helpful clues for intra-sentential RE~\cite{miwa-bansal2016}. We thus use labelled syntactic dependency edges between the words of each sentence, by treating each syntactic dependency label as a different edge type.

\noindent \textbf{Coreference edge:} As coreference is an important indicator of local and non-local dependencies~\cite{ma2016unsupervised}, we connect co-referring phrases in a document using coreference type edges.

\noindent \textbf{Adjacent sentence edge:} We connect the syntactic root of a sentence with the roots of the previous and next sentences with adjacent sentence type edges~\cite{peng2017} for non-local dependencies between neighbouring sentences. 

\noindent \textbf{Adjacent word edge:} In order to keep sequential information among the words of a sentence, we connect each word with its previous and next words with adjacent word type edges.

\noindent \textbf{Self-node edge:} GCNN learns a node representation based solely on its neighbour nodes and their edge types. 
Hence, to include the node information itself into the representation, we form self-node type edges on all the nodes of the graph.

\subsection{GCNN Layer} 
\label{sec:graph_cnn}

We compute the representation of each input word $i$ by applying GCNN~\cite{kipf2016semi,defferrard2016} on the constructed document graph. GCNN is an advanced version of CNN for graph encoding that learns semantic representations for the graph nodes, while preserving its structural information. 
In order to learn edge type-specific representations, we use a labelled edge GCNN, which keeps separate parameters for each edge type~\cite{Shikhar2018}. 
The GCNN iteratively updates the representation of each input word $i$ as follows:
\begin{equation*}
\mathbf{x}_{i}^{k+1} = f \left( \sum_{u \in \nu(i)} \left( \mathbf{W}^{k}_{l(i,u)} \; \mathbf{x}_{u}^{k} + \mathbf{b}^k_{l(i,u)}  \right) \right),
\end{equation*}
where $\mathbf{x}_i^{k+1}$ is the $i$-th word representation resulted from the $k$-th GCNN block, $\nu(i)$ is a set of neighbouring nodes to $i$, $\mathbf{W}^{k}_{l(i,u)}$ and $\mathbf{b}^k_{l(i,u)}$ are the parameters of the $k$-th block for edge type $l$ between nodes $i$ and $u$. 
We stack $K$ GCNN blocks to accumulate information from distant neighbouring nodes and use edge-wise gating to control information from neighbouring nodes. 

Similar to \citet{Marcheggiani2017}, we maintain separate parameters for each edge direction. 
We, however, tune the number of model parameters by keeping separate parameters only for the top-$N$ types and using the same parameters for all the remaining edge types, named ``rare'' type edges. 
This can avoid possible overfitting due to over-parameterisation for different edge types.

\subsection{MIL-based Relation Classification}  
\label{sec:bi-affine}
Since each target entity can have multiple mentions in a document, 
we employ a multi-instance learning (MIL)-based classification scheme to aggregate the predictions of all target mention pairs using bi-affine pairwise scoring~\cite{Patrick2018}.
As shown in Figure~\ref{fig:model}, each word $i$ is firstly projected into two separate latent spaces using two-layered feed-forward neural networks (FFNN), which correspond to the first (head) or second (tail) argument of the target pair.
\begin{align*}
	\mathbf{x}_i^{head} &= \mathbf{W}^{(1)}_{head} \left( \relu \left(\mathbf{W}^{(0)}_{head} \; \mathbf{x}_i^{K} \right) \right), \\
	\mathbf{x}_i^{tail} &= \mathbf{W}^{(1)}_{tail} \left( \relu \left( \mathbf{W}^{(0)}_{tail} \; \mathbf{x}_i^{K} \right) \right), \nonumber
\end{align*}
where $\mathbf{x}_i^K$ corresponds to the representation of the $i$-th word after $|K|$ blocks of GCNN encoding,
$\mathbf{W}^{(0)}$, $\mathbf{W}^{(1)}$ are the parameters of two FFNNs for head and tail respectively
and $\mathbf{x}_i^{head}$, $\mathbf{x}_i^{tail} \in \mathbb{R}^d$ are the resulted head/tail representations for the $i$-th word. 

Then, mention-level pairwise confidence scores are generated by a bi-affine layer and aggregated to obtain the entity-level pairwise score. 
\begin{multline*}
    scores(e^{head}, e^{tail}) = \\
    \log \sum_{\substack{i \in E^{head},\  j \in E^{tail}}} \exp{ \left( \left( \mathbf{x}_i^{head} \; \mathbf{R} \right) \mathbf{x}_j^{tail} \right) },
\end{multline*}
where, $\mathbf{R} \in \mathbb{R}^{d\times r\times d}$ is a learned bi-affine tensor with $r$ the number of relation categories,
and $E^{head}$, $E^{tail}$ denote a set of mentions for entities $e^{head}$ and $e^{tail}$ respectively.

\section{Experimental Settings}
We first briefly describe the datasets where the proposed model is evaluated along with their pre-processing. We then introduce the baseline models we use for comparison. Finally, we show the training settings. 

\subsection{Data Sets}
We evaluated our model on two biochemistry datasets.\\
\textbf{Chemical-Disease Relations dataset (CDR):}
The CDR dataset is a document-level, inter-sentence relation extraction dataset developed for the \textit{BioCreative V} challenge~\cite{biocreative2015overview}.

\noindent \textbf{CHemical Reactions dataset (CHR):}
We created a document-level dataset with relations between chemicals using distant supervision.
Firstly, we used the back-end of the semantic faceted search engine Thalia\footnote{\url{http://www.nactem.ac.uk/Thalia/}}~\cite{Thalia2018} to obtain abstracts annotated with several biomedical named entities from PubMed. 
We selected chemical compounds from the annotated entities and aligned them with the graph database \textit{Biochem4j}~\cite{Biochem4j}. \textit{Biochem4j} is a freely available database that integrates several resources such as UniProt, KEGG and NCBI Taxonomy\footnote{\url{http://biochem4j.org}}. 
If two chemical entities have a relation in \textit{Biochem4j}, we consider them as positive instances in the dataset, otherwise as negative.

\subsection{Data Pre-processing}

Table~\ref{tab:data_stats} shows the statistics for CDR and CHR datasets. 
For both datasets, the annotated entities can have more than one associated Knowledge Base (KB) ID. If there is at least one common KB ID between mentions then we considered all these mentions to belong to the same entity. This technique results in less negative pairs. 
We ignored entities that were not grounded to a known KB ID and removed relations between the same entity (self-relations).  
For the CDR dataset, we performed hypernym filtering similar to \citet{gu2017} and \citet{Patrick2018}.
In the CHR dataset, both directions were generated for each candidate chemical pair as chemicals can be either a reactant (first argument) or a product (second argument) in an interaction.

\begin{table}[t!]
	\centering
	\scalebox{0.85}{
	\begin{tabular}{llrrr} 
		\hline
		\textbf{Data} & \textbf{Count} & \textbf{Train} & \textbf{Dev.} & \textbf{Test}  \\
		\hline 
		\multirow{3}{*}{CDR} 
		    & \# Articles     & 500 & 500 & 500 \\ 
		    & \# Positive pairs  & 1,038  & 1,012 & 1,066 \\ 
		    & \# Negative pairs  & 4,198  & 4,069 & 4,119 \\
		\hline
		\multirow{3}{*}{CHR} 
		    & \# Articles     & 7,298 & 1,182 & 3,614 \\ 
		    & \# Positive pairs  & 19,643 & 3,185 & 9,578 \\              
		    & \# Negative pairs  & 69,843 & 11,466 & 33,339 \\ 
		    \hline     
	\end{tabular} 
	}
	\caption{Statistics of the CDR and CHR datasets.}
	\label{tab:data_stats}
\end{table}

We processed the datasets using the GENIA Sentence Splitter\footnote{\url{http://www.nactem.ac.uk/y-matsu/geniass/}} and GENIA tagger~\cite{tsuruoka2005developing} for sentence splitting and word tokenisation, respectively. Syntactic dependencies were obtained using the Enju syntactic parser~\cite{miyao-tsujii-2008-feature} with predicate-argument structures. Coreference type edges were constructed using the Stanford CoreNLP software~\cite{corenlp:2014}. 

\subsection{Baseline Models}
For the CDR dataset, we compare with five state-of-the-art models: SVM~\cite{xu-EtAl:2016}, ensemble of feature-based and neural-based models~\cite{zhou2016cdr}, CNN and Maximum Entropy~\cite{gu2017}, Piece-wise CNN~\cite{Li2018} and Transformer~\cite{Patrick2018}.
We additionally prepare and evaluate the following models: 
\textbf{CNN-RE}, a re-implementation from~\citet{kim2014} and \citet{zhou2016cdr} and \textbf{RNN-RE}, a re-implementation from \citet{sunil2017}.
In all models we use bi-affine pairwise scoring to detect relations.

\subsection{Model Training}
\label{sec:model_training}
We used 100-dimentional word embeddings trained on PubMed with GloVe~\cite{pennington2014glove,th2015evaluating}. Unlike \citet{Patrick2018}, we used the pre-trained word embeddings in place of sub-word embeddings to align with our word graphs.
Due to the size of the CDR dataset, we merged the training and development sets to train the models, similarly to \citet{Jun2016} and \citet{gu2017}.
We report the performance as the average of five runs with different parameter initialisation seeds in terms of precision (P), recall (R) and F1-score. 
We used the frequencies of the edge types in the training set to choose the top-$N$ edges in Section~\ref{sec:graph_cnn}. 
We refer to  the supplementary materials for the details of the training and hyper-parameter settings.

\section{Results}
\label{sec:exp}

We show the results of our model for the CDR and CHR datasets in Table~\ref{tab:results}.
We report the performance of state-of-the-art models without any additional enhancements, such as joint training with NER, model ensembling and heuristic rules, to avoid any effects from the enhancements in the comparison. 
We observe that the GCNN outperforms the baseline models (CNN-RE/RNN-RE) in both datasets. 
However, in the CDR dataset, the performance of GCNN is $1.6$ percentage points lower than the best performing system of \cite{gu2017}. In fact, \citet{gu2017} incorporates two separate neural and feature-based models for intra- and inter-sentence pairs, respectively, whereas we utilize a single model for both pairs.
Additionally, GCNN performs comparably to the second state-of-the-art neural model~\citet{Li2018}, which requires a two-step process for mention aggregation unlike our unified approach.

\begin{table}[t!]
    \centering
    \scalebox{0.85}{
    \begin{tabular}{llccc} 
        \hline
        \textbf{Data} & \textbf{Model} & \textbf{P} (\%) & \textbf{R} (\%) & \textbf{F1} (\%)  \\
        \hline 
        \multirow{8}{*}{CDR} 
        & \citet{Jun2016}	           & 59.6 & 44.0 & 50.7  \\  
        & \citet{zhou2016cdr}  	       & 64.8 & 49.2 & 56.0  \\  
        & \citet{gu2017}               & 60.9 & 59.5 & 60.2  \\ 
        & \citet{Li2018}       		   & 55.1 & 63.6 & 59.1  \\  
        & \citet{Patrick2018}          & 49.9 & 63.8 & 55.5  \\ 
        
        & CNN-RE                       & 51.5 & 65.7 & 57.7  \\ 
        & RNN-RE                       & 52.6 & 62.9 & 57.3  \\
        & \textbf{GCNN} 			   & 52.8 & 66.0 & 58.6 \\
        \hline 
        
        \multirow{3}{*}{CHR} 
        & CNN-RE                       & 81.2 & 87.3 & 84.1  \\
        & RNN-RE                       & 83.0 & 90.1 & 86.4  \\ 
        & \textbf{GCNN}                & 84.7 & 90.5 & \textbf{87.5} \\
        \hline
    \end{tabular}
    }
    \caption{Performance on the CDR and CHR test sets in comparison with the state-of-the-art.}
    \label{tab:results}
\end{table}

\begin{figure}[t!]
    \centering
    \includegraphics[width=0.48\textwidth]{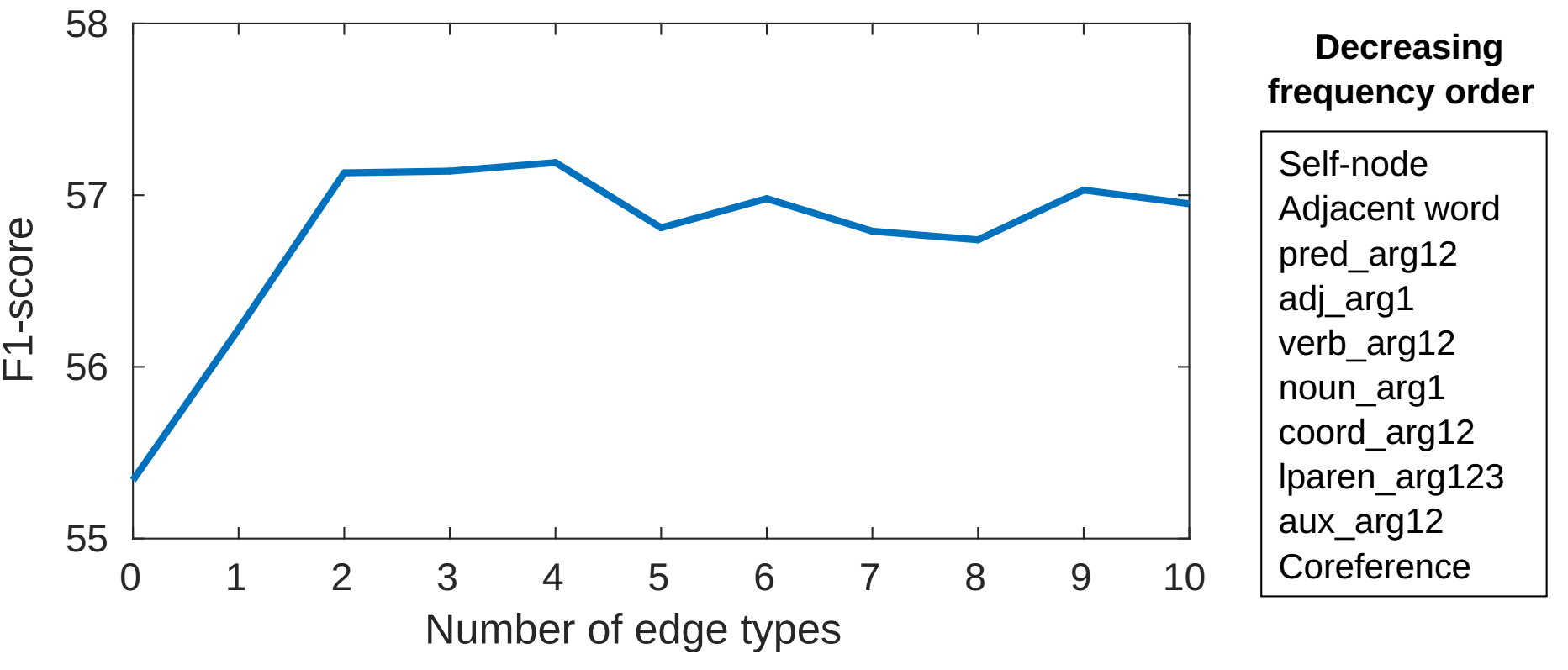}
    \caption{Performance of GCNN model on the CDR development set when using the top-$N$ most frequent edge types and consider the rest as a single ``rare'' type.} 
    \label{fig:edgeset}
\end{figure}

Figure~\ref{fig:edgeset} illustrates the performance of our model on the CDR development set when using a varying number of most frequent edge types $N$.
While tuning $N$, we observed that the best performance was obtained for top-$4$ edge types, but it slightly deteriorated with more. We chose the top-$4$ edge types in other experiments.

We perform ablation analysis on the CDR dataset by separating the development set to intra- and inter-sentence pairs (approximately 70\% and 30\% of pairs, respectively). Table~\ref{tab:ablation_res} shows the performance when removing an edge category at a time. 
In general, all dependency types have positive effects on inter-sentence RE and the overall performance, although self-node and adjacent sentence edges slightly harm the performance of intra-sentence relations. Additionally, coreference does not affect intra-sentence pairs.

\begin{table}[t!]
	\centering
	\scalebox{0.85}{
	\begin{tabular}{lccc} 
		\hline
		\textbf{Model} &\textbf{Overall} & \textbf{Intra} & \textbf{Inter} \\ \hline 
		\textbf{GCNN} (best)      & \textbf{57.19} & 63.43 & \textbf{36.90} \\ 
		\ \ $-$ Adjacent word        & 55.75     & 62.53 & 35.61 \\ 
		\ \ $-$ Syntactic dependency & 56.12     & 62.89 & 34.75 \\ 
		\ \ $-$ Coreference          & 56.44     & 63.27 & 35.65 \\  
	    \ \ $-$ Self-node            & 56.85     & 63.84 & 33.20 \\  
	    \ \ $-$ Adjacent sentence    & 57.00     & \textbf{63.99} & 35.20 \\ 
	    \hline 
	\end{tabular}
	}
	\caption{Ablation analysis on the CDR development set, in terms of F1-score (\%), for intra- (Intra) and inter-sentence (Inter) pairs.}
	\label{tab:ablation_res}
\end{table}

\section{Related Work}
\label{sec:rel_work}

Inter-sentence RE is a recently introduced task. \citet{peng2017} and \citet{song2018n} used graph-based LSTM networks for $n$-ary RE in multiple sentences for protein-drug-disease associations. They restricted the relation candidates in up to two-span sentences. \citet{Patrick2018} considered multi-instance learning for document-level RE. Our work is different from \citet{Patrick2018} in that we replace Transformer with a GCNN model for full-abstract encoding using non-local dependencies such as entity coreference.

GCNN was firstly proposed by \citet{kipf2016semi} and applied on citation networks and knowledge graph datasets. 
It was later used for semantic role labelling~\cite{Marcheggiani2017}, multi-document summarization~\cite{Yasunaga2017} and temporal relation extraction~\cite{Shikhar2018}. \citet{zhang2018graph} used a GCNN on a dependency tree for intra-sentence RE. 
Unlike previous work, we introduced a GCNN on a document-level graph, with both intra- and inter-sentence dependencies for inter-sentence RE.

\section{Conclusion}
\label{sec:conc}
We proposed a novel graph-based method for inter-sentence RE using a labelled edge GCNN model on a document-level graph. The graph is constructed with words as nodes and multiple intra- and inter-sentence dependencies between them as edges. 
A GCNN model is employed to encode the graph structure and MIL is incorporated to aggregate the multiple mention-level pairs .
We show that our method achieves comparable performance to the state-of-the-art neural models on two biochemistry datasets. We tuned the number of labelled edges to maintain the number of parameters in the labelled edge GCNN. Analysis showed that all edge types are effective for inter-sentence RE.

Although the model is applied to biochemistry corpora for inter-sentence RE, our method is also applicable to other relation extraction tasks.
As future work, we plan to incorporate joint named entity recognition training as well as sub-word embeddings in order to further improve the performance of the proposed model.

\section*{Acknowledgments}
This research was supported with funding from BBSRC, Enriching Metabolic PATHwaY models with evidence from the literature (EMPATHY) [Grant ID: BB/M006891/1] and AIRC/AIST. Results were obtained from a project commissioned by the New Energy and Industrial Technology Development Organization (NEDO).

\bibliographystyle{acl_natbib}
\bibliography{biblio}

\begin{thebibliography}{35}
\expandafter\ifx\csname natexlab\endcsname\relax\def\natexlab#1{#1}\fi

\bibitem[{Culotta and Sorensen(2004)}]{culotta2004dependency}
Aron Culotta and Jeffrey Sorensen. 2004.
\newblock Dependency tree kernels for relation extraction.
\newblock In \emph{Proceedings of Annual Meeting on Association for
  Computational Linguistics}, pages 423--430. Association for Computational
  Linguistics.

\bibitem[{Defferrard et~al.(2016)Defferrard, Bresson, and
  Vandergheynst}]{defferrard2016}
Micha{\"e}l Defferrard, Xavier Bresson, and Pierre Vandergheynst. 2016.
\newblock Convolutional neural networks on graphs with fast localized spectral
  filtering.
\newblock In \emph{Advances in Neural Information Processing Systems}, pages
  3844--3852.

\bibitem[{Gu et~al.(2017)Gu, Sun, Qian, and Zhou}]{gu2017}
Jinghang Gu, Fuqing Sun, Longhua Qian, and Guodong Zhou. 2017.
\newblock Chemical-induced disease relation extraction via convolutional neural
  network.
\newblock \emph{Database}, 2017:1--12.

\bibitem[{Kim(2014)}]{kim2014}
Yoon Kim. 2014.
\newblock Convolutional neural networks for sentence classification.
\newblock In \emph{Proceedings of Conference on Empirical Methods in Natural
  Language Processing}, pages 1746--1751. Association for Computational
  Linguistics.

\bibitem[{Kingma and Ba(2015)}]{kingma2014adam}
Diederik~P Kingma and Jimmy Ba. 2015.
\newblock Adam: A method for stochastic optimization.
\newblock In \emph{Proceedings of the International Conference on Learning
  Representations}.

\bibitem[{Kipf and Welling(2017)}]{kipf2016semi}
Thomas~N Kipf and Max Welling. 2017.
\newblock Semi-supervised classification with graph convolutional networks.
\newblock In \emph{International Conference on Learning Representations}.

\bibitem[{Li et~al.(2018)Li, Yang, Chen, Tang, Wang, and Yan}]{Li2018}
Haodi Li, Ming Yang, Qingcai Chen, Buzhou Tang, Xiaolong Wang, and Jun Yan.
  2018.
\newblock Chemical-induced disease extraction via recurrent piecewise
  convolutional neural networks.
\newblock \emph{BMC Medical Informatics and Decision Making}, 18(2):60.

\bibitem[{Lin et~al.(2016)Lin, Shen, Liu, Luan, and Sun}]{lin2016neural}
Yankai Lin, Shiqi Shen, Zhiyuan Liu, Huanbo Luan, and Maosong Sun. 2016.
\newblock Neural relation extraction with selective attention over instances.
\newblock In \emph{Proceedings of Annual Meeting of the Association for
  Computational Linguistics}, volume~1, pages 2124--2133. Association for
  Computational Linguistics.

\bibitem[{Liu et~al.(2015)Liu, Wei, Li, Ji, Zhou, and Wang}]{liu2015dependency}
Yang Liu, Furu Wei, Sujian Li, Heng Ji, Ming Zhou, and Houfeng Wang. 2015.
\newblock A dependency-based neural network for relation classification.
\newblock In \emph{Proceedings of the 53rd Annual Meeting of the Association
  for Computational Linguistics and the 7th International Joint Conference on
  Natural Language Processing}, pages 285--290. Association for Computational
  Linguistics.

\bibitem[{Ma et~al.(2016)Ma, Liu, and Hovy}]{ma2016unsupervised}
Xuezhe Ma, Zhengzhong Liu, and Eduard Hovy. 2016.
\newblock Unsupervised ranking model for entity coreference resolution.
\newblock In \emph{Proceedings of the Conference of the North American Chapter
  of the Association for Computational Linguistics: Human Language
  Technologies}, pages 1012--1018. Association for Computational Linguistics.

\bibitem[{Manning et~al.(2014)Manning, Surdeanu, Bauer, Finkel, Bethard, and
  McClosky}]{corenlp:2014}
Christopher~D Manning, Mihai Surdeanu, John Bauer, Jenny Finkel, Steven
  Bethard, and David McClosky. 2014.
\newblock The {Stanford} {CoreNLP} natural language processing toolkit.
\newblock In \emph{Proceedings of 52nd Annual Meeting of the Association for
  Computational Linguistics: System Demonstrations}, pages 55--60. Association
  for Computational Linguistics.

\bibitem[{Marcheggiani and Titov(2017)}]{Marcheggiani2017}
Diego Marcheggiani and Ivan Titov. 2017.
\newblock Encoding sentences with graph convolutional networks for semantic
  role labeling.
\newblock In \emph{Proceedings of Conference on Empirical Methods in Natural
  Language Processing}, pages 1506--1515. Association for Computational
  Linguistics.

\bibitem[{Miwa and Bansal(2016)}]{miwa-bansal2016}
Makoto Miwa and Mohit Bansal. 2016.
\newblock End-to-end relation extraction using lstms on sequences and tree
  structures.
\newblock In \emph{Proceedings of the Annual Meeting of the Association for
  Computational Linguistics}, pages 1105--1116. Association for Computational
  Linguistics.

\bibitem[{Miyao and Tsujii(2008)}]{miyao-tsujii-2008-feature}
Yusuke Miyao and Jun{'}ichi Tsujii. 2008.
\newblock Feature forest models for probabilistic {HPSG} parsing.
\newblock \emph{Computational Linguistics}, 34(1):35--80.

\bibitem[{Peng et~al.(2017)Peng, Poon, Quirk, Toutanova, and Yih}]{peng2017}
Nanyun Peng, Hoifung Poon, Chris Quirk, Kristina Toutanova, and Wen-tau Yih.
  2017.
\newblock Cross-sentence n-ary relation extraction with {Graph LSTMs}.
\newblock \emph{Transactions of the Association for Computational Linguistics},
  5:101--115.

\bibitem[{Pennington et~al.(2014)Pennington, Socher, and
  Manning}]{pennington2014glove}
Jeffrey Pennington, Richard Socher, and Christopher~D Manning. 2014.
\newblock {GloVe}: Global vectors for word representation.
\newblock In \emph{Proceedings of the Conference on Empirical Methods in
  Natural Language Processing}, pages 1532--1543. Association for Computational
  Linguistics.

\bibitem[{Riedel et~al.(2010)Riedel, Yao, and McCallum}]{riedel2010}
Sebastian Riedel, Limin Yao, and Andrew McCallum. 2010.
\newblock Modeling relations and their mentions without labeled text.
\newblock In \emph{Joint European Conference on Machine Learning and Knowledge
  Discovery in Databases}, pages 148--163. Springer.

\bibitem[{Sahu and Anand(2018)}]{sunil2017}
Sunil~Kumar Sahu and Ashish Anand. 2018.
\newblock Drug-drug interaction extraction from biomedical texts using long
  short-term memory network.
\newblock \emph{Journal of Biomedical Informatics}, 86:15 -- 24.

\bibitem[{dos Santos et~al.(2015)dos Santos, Xiang, and Zhou}]{dos2015}
C{\i}cero~Nogueira dos Santos, Bing Xiang, and Bowen Zhou. 2015.
\newblock Classifying relations by ranking with convolutional neural networks.
\newblock In \emph{Proceedings of the Annual Meeting of the Association for
  Computational Linguistics and the International Joint Conference on Natural
  Language Processing}, pages 626--634. Association for Computational
  Linguistics.

\bibitem[{Song et~al.(2018)Song, Zhang, Wang, and Gildea}]{song2018n}
Linfeng Song, Yue Zhang, Zhiguo Wang, and Daniel Gildea. 2018.
\newblock N-ary relation extraction using graph-state {LSTM}.
\newblock In \emph{Proceedings of Conference on Empirical Methods in Natural
  Language Processing}, pages 2226--2235. Association for Computational
  Linguistics.

\bibitem[{Soto et~al.(2018)Soto, Przyby{\l}a, and Ananiadou}]{Thalia2018}
Axel~J Soto, Piotr Przyby{\l}a, and Sophia Ananiadou. 2018.
\newblock Thalia: Semantic search engine for biomedical abstracts.
\newblock \emph{Bioinformatics}.

\bibitem[{Surdeanu et~al.(2012)Surdeanu, Tibshirani, Nallapati, and
  Manning}]{surdeanu2012}
Mihai Surdeanu, Julie Tibshirani, Ramesh Nallapati, and Christopher~D Manning.
  2012.
\newblock Multi-instance multi-label learning for relation extraction.
\newblock In \emph{Proceedings of the Joint Conference on Empirical Methods in
  Natural Language Processing and Computational Natural Language Learning},
  pages 455--465. Association for Computational Linguistics.

\bibitem[{Swainston et~al.(2017)Swainston, Batista-Navarro, Carbonell, Dobson,
  Dunstan, Jervis, Vinaixa, Williams, Ananiadou, Faulon et~al.}]{Biochem4j}
Neil Swainston, Riza Batista-Navarro, Pablo Carbonell, Paul~D Dobson, Mark
  Dunstan, Adrian~J Jervis, Maria Vinaixa, Alan~R Williams, Sophia Ananiadou,
  Jean-Loup Faulon, et~al. 2017.
\newblock biochem4j: Integrated and extensible biochemical knowledge through
  graph databases.
\newblock \emph{PloS one}, 12(7):e0179130.

\bibitem[{TH et~al.(2015)TH, Sahu, and Anand}]{th2015evaluating}
MUNEEB TH, Sunil Sahu, and Ashish Anand. 2015.
\newblock Evaluating distributed word representations for capturing semantics
  of biomedical concepts.
\newblock In \emph{Proceedings of BioNLP}, pages 158--163. Association for
  Computational Linguistics.

\bibitem[{Tsuruoka et~al.(2005)Tsuruoka, Tateishi, Kim, Ohta, McNaught,
  Ananiadou, and Tsujii}]{tsuruoka2005developing}
Yoshimasa Tsuruoka, Yuka Tateishi, Jin-Dong Kim, Tomoko Ohta, John McNaught,
  Sophia Ananiadou, and Jun’ichi Tsujii. 2005.
\newblock Developing a robust part-of-speech tagger for biomedical text.
\newblock In \emph{Panhellenic Conference on Informatics}, pages 382--392.
  Springer.

\bibitem[{Vashishth et~al.(2018)Vashishth, Dasgupta, Ray, and
  Talukdar}]{Shikhar2018}
Shikhar Vashishth, Shib~Sankar Dasgupta, Swayambhu~Nath Ray, and Partha
  Talukdar. 2018.
\newblock Dating documents using graph convolution networks.
\newblock In \emph{Proceedings of the Annual Meeting of the Association for
  Computational Linguistics}, pages 1605--1615. Association for Computational
  Linguistics.

\bibitem[{Verga et~al.(2018)Verga, Strubell, and McCallum}]{Patrick2018}
Patrick Verga, Emma Strubell, and Andrew McCallum. 2018.
\newblock Simultaneously self-attending to all mentions for full-abstract
  biological relation extraction.
\newblock In \emph{Proceedings of Conference of the North American Chapter of
  the Association for Computational Linguistics: Human Language Technologies},
  pages 872--884. Association for Computational Linguistics.

\bibitem[{Wei et~al.(2015)Wei, Peng, Leaman, Davis, Mattingly, Li, Wiegers, and
  Lu}]{biocreative2015overview}
Chih-Hsuan Wei, Yifan Peng, Robert Leaman, Allan~Peter Davis, Carolyn~J
  Mattingly, Jiao Li, Thomas~C Wiegers, and Zhiyong Lu. 2015.
\newblock {Overview of the BioCreative V chemical disease relation (CDR) task}.
\newblock In \emph{Proceedings of the fifth BioCreative challenge evaluation
  workshop}, pages 154--166.

\bibitem[{Xu et~al.(2016{\natexlab{a}})Xu, Wu, Zhang, Wang, Lee, and
  Xu}]{Jun2016}
Jun Xu, Yonghui Wu, Yaoyun Zhang, Jingqi Wang, Hee-Jin Lee, and Hua Xu.
  2016{\natexlab{a}}.
\newblock {CD-REST}: a system for extracting chemical-induced disease relation
  in literature.
\newblock \emph{Database}, 2016:1--10.

\bibitem[{Xu et~al.(2016{\natexlab{b}})Xu, Reddy, Feng, Huang, and
  Zhao}]{xu-EtAl:2016}
Kun Xu, Siva Reddy, Yansong Feng, Songfang Huang, and Dongyan Zhao.
  2016{\natexlab{b}}.
\newblock Question answering on freebase via relation extraction and textual
  evidence.
\newblock In \emph{Proceedings of the 54th Annual Meeting of the Association
  for Computational Linguistics}, pages 2326--2336. Association for
  Computational Linguistics.

\bibitem[{Yasunaga et~al.(2017)Yasunaga, Zhang, Meelu, Pareek, Srinivasan, and
  Radev}]{Yasunaga2017}
Michihiro Yasunaga, Rui Zhang, Kshitijh Meelu, Ayush Pareek, Krishnan
  Srinivasan, and Dragomir Radev. 2017.
\newblock Graph-based neural multi-document summarization.
\newblock In \emph{Proceedings of the 21st Conference on Computational Natural
  Language Learning}, pages 452--462. Association for Computational
  Linguistics.

\bibitem[{Zeng et~al.(2014)Zeng, Liu, Lai, Zhou, and Zhao}]{Zeng14}
Daojian Zeng, Kang Liu, Siwei Lai, Guangyou Zhou, and Jun Zhao. 2014.
\newblock {Relation Classification via Convolutional Deep Neural Network}.
\newblock In \emph{Proceedings of the International Conference on Computational
  Linguistics}, pages 2335--2344.

\bibitem[{Zhang et~al.(2018)Zhang, Qi, and Manning}]{zhang2018graph}
Yuhao Zhang, Peng Qi, and Christopher~D Manning. 2018.
\newblock Graph convolution over pruned dependency trees improves relation
  extraction.
\newblock In \emph{Proceedings of the 2018 Conference on Empirical Methods in
  Natural Language Processing}, pages 2205--2215. Association for Computational
  Linguistics.

\bibitem[{Zhou et~al.(2016{\natexlab{a}})Zhou, Deng, Chen, Yang, Jia, and
  Huang}]{zhou2016cdr}
Huiwei Zhou, Huijie Deng, Long Chen, Yunlong Yang, Chen Jia, and Degen Huang.
  2016{\natexlab{a}}.
\newblock Exploiting syntactic and semantics information for chemical–disease
  relation extraction.
\newblock \emph{Database}, 2016:1--12.

\bibitem[{Zhou et~al.(2016{\natexlab{b}})Zhou, Shi, Tian, Qi, Li, Hao, and
  Xu}]{zhou2016}
Peng Zhou, Wei Shi, Jun Tian, Zhenyu Qi, Bingchen Li, Hongwei Hao, and Bo~Xu.
  2016{\natexlab{b}}.
\newblock {Attention-Based Bidirectional Long Short-Term Memory Networks for
  Relation Classification}.
\newblock In \emph{Proceedings of the Annual Meeting of the Association for
  Computational Linguistics}, pages 207--212. Association for Computational
  Linguistics.

\end{thebibliography}

\newpage

\appendix

\section{Training and Hyper-parameter Settings}
We implemented all models using \textit{Tensorflow}\footnote{\url{https://www.tensorflow.org}}.
The development set was used for hyper-parameter tuning.
For all models, parameters were optimised using the Adam optimisation algorithm with exponential moving average~\cite{kingma2014adam}, learning rate of $0.0005$, learning rate decay of $0.75$ and gradient clipping $10$. We used early stopping with patience equal to $5$ epochs in order to determine the best training epoch. For other hyper-parameters, we performed a non-exhaustive hyper-parameter search based on the development set.
We used the same hyper-parameters of both CDR and CHR datasets. The best hyper-parameter values are shown in Table~\ref{tab:hyperparameters}. 

\begin{table}[ht!]
	\centering
		\scalebox{0.9}{
		\begin{tabular}{lr} 
			\hline
			\textbf{Hyper-parameter}      & \textbf{Value}  \\  \hline
			Batch size 		                        & 32           \\  
			Learning rate                           & 5 $\cdot$ 10$^{-3}$  \\ 
			Word dimension                          & 100    \\  
			Position dimension                      & 20     \\  
			GCNN dimension                          & 140    \\  
			Number of GCNN blocks ($K$)             & 2      \\  
			MIL feed-forward layer dimension        & 140     \\ 
			Dropout rate (input layer)              & 0.1     \\  
			Dropout rate (GCNN layer)               & 0.05    \\ 
			Dropout rate (MIL feed-forward layer)   & 0.05    \\ 
			Residual connection on GCNN layer       & yes     \\ \hline 
		\end{tabular}
		}
	\caption{Best performing hyper-parameters used in the proposed model.}
	\label{tab:hyperparameters}
\end{table}

\end{document}